\documentclass{article}
\usepackage{amssymb}
\usepackage{xcolor}
\usepackage{wrapfig}
\usepackage[final]{corl_2025} 

\usepackage{graphics,graphicx,float,caption,subcaption,booktabs,xcolor,multirow,array,color,ifthen,tabu,colortbl,dblfloatfix,url,xparse,mathtools,patchcmd,algorithm,algorithmicx,algpseudocode,amssymb,xspace,nicefrac,microtype,amsmath,amsfonts,bm,ragged2e,tikz,stackengine,etoolbox,xpatch,enumerate,enumitem,setspace,tabularx,makecell,cuted,titlesec,enumitem,wrapfig,tcolorbox,verbatim,footmisc,arydshln,duckuments,framed,hyperref,dsfont,siunitx}

\newcommand{\ours}{LocoFormer\xspace}

\title{LocoFormer: Generalist Locomotion via\\Long-context Adaptation}

\author{ \hspace{2.5em}   Min Liu \quad Deepak Pathak \quad Ananye Agarwal \\
\small Skild AI
}

\begin{document}
\maketitle
\vspace{-15pt}
\begin{figure*}[h]  
    \centering
    \includegraphics[width=1.0\textwidth]{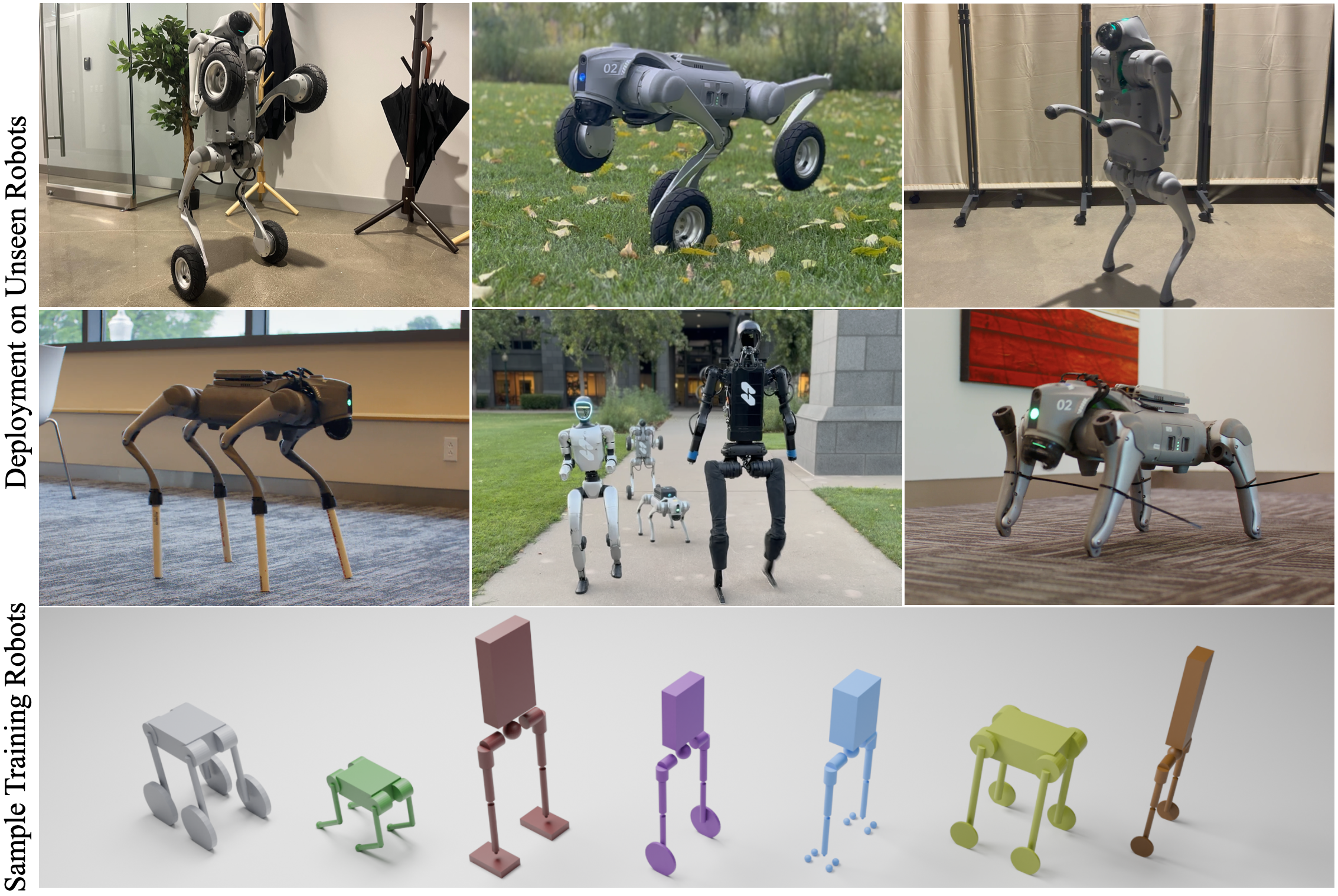}
    \vspace{-12pt}
    \caption{\small We propose \ours: \textbf{\textit{a generalist omni-bodied locomotion model}} that transfers \textit{zero-shot} to any kind of robot body, whether completely different form-factor or bodily changes due to damaged limbs, without even training on them (top two rows). The key innovation is \textit{long-term multi-episodic adaptation} trained with massively large-scale reinforcement learning -- this allows generalization to a variety of real robots by just training on randomly generated procedural robots (bottom row).
    We see emergent adaptation behavior across very long contexts and even few-shot multi-episodic adaptation across a handful of trials where \ours uses suboptimal initial rollouts to improve in subsequent ones. Videos are at \href{https://generalist-locomotion.github.io}{generalist-locomotion.github.io}.
    }
    \label{fig:teaser}
\end{figure*}
\vspace{-1em}
\begin{abstract}
Modern locomotion controllers are manually tuned for specific embodiments. We present \textit{\ours}, a generalist omni-bodied locomotion model that can control previously unseen legged and wheeled robots, even without precise knowledge of their kinematics. \ours is able to adapt to changes in morphology and dynamics at test time. We find that two key choices enable adaptation. First, we train massive scale RL on procedurally generated robots with aggressive domain randomization. Second, in contrast to previous policies that are myopic with short context lengths, we extend context by orders of magnitude to span episode boundaries. We deploy the same \ours to varied robots and show robust control even with large disturbances such as weight change and motor failures. In extreme scenarios, we see emergent adaptation across episodes, \ours learns from falls in early episodes to improve control strategies in later ones. We believe that this simple, yet general recipe can be used to train foundation models for other robotic skills in the future. Videos at \href{https://generalist-locomotion.github.io}{generalist-locomotion.github.io}.

\end{abstract}

\clearpage

\section{Introduction}
\begin{wrapfigure}{r}{0.45\linewidth}
    \vspace{-1em}
    \centering
    \includegraphics[width=\linewidth]{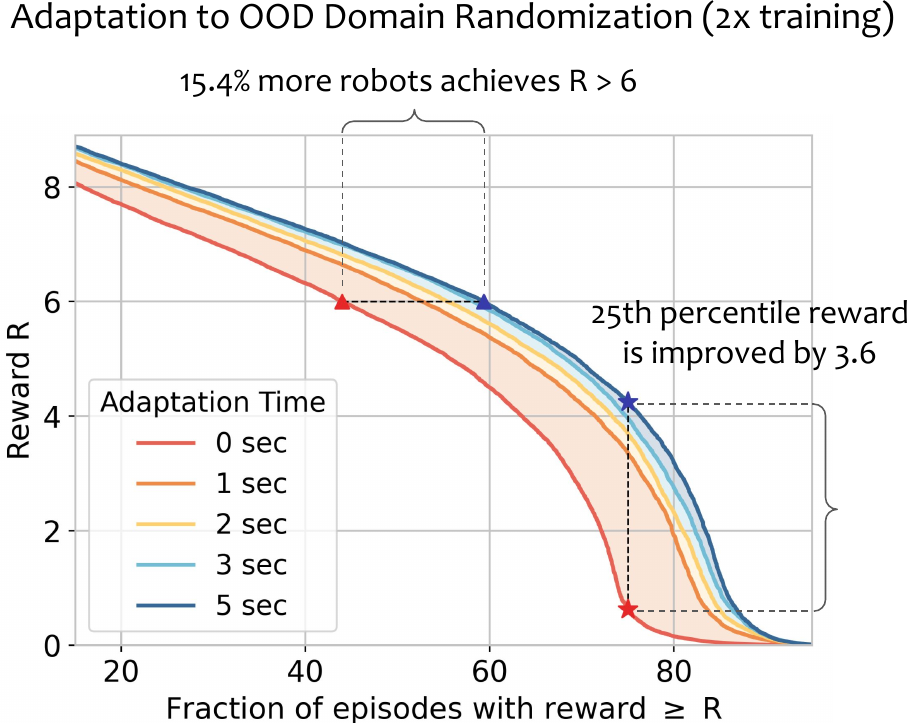}
    \caption{\small We test on OOD scenario by doubling the ranges of domain randomization. We see that \ours uses multiple seconds of history to improve performance. In contrast, previous policies are myopic and fail to improve beyond few hundreds of milliseconds. Figure design from \cite{team2023human}}
    \label{fig:adapt_motivation}
    \vspace{-1.25em}
\end{wrapfigure}
\vspace{-2em}
Humans and animals are remarkably flexible at control. In the case of locomotion, they exhibit highly adaptive behaviors, such as learning to walk shortly after being born \citep{muir2000early}, or adapting to limb amputations \cite{GOLDNER2015192}. Even humans at birth possess reflexes such as alternating stepping that promote the learning of robust locomotion \cite{thelen1982newborn}. This is in stark contrast to robotic locomotion today, which is typically learned \textit{tabula rasa} with reinforcement learning and tuned for a specific embodiment \cite{kumar2021rma, lee2020learning, nahrendra2023dreamwaq, margolis2023walk, 10530429, smith2023learning}. Although such policies adapt to some randomization, this is very narrow and myopic over a timescale of a few hundred milliseconds \cite{radosavovic2024real, agarwal2023legged, kumar2021rma}. This design choice favors in-distribution performance over generality, since myopic policies fail to adapt to wider task distributions, leading to lower performance. These resulting controllers therefore do not generalize to new embodiments and fail catastrophically in the case of large changes such as burnt out motors, broken legs, or modeling errors. This often means that successful sim2real transfer requires painstaking system identification for each robot. Further, a narrow task distribution is unlikely to promote the learning of general and re-usable strategies.

In this paper, We present \textit{\ours}, a generalist omni-bodied locomotion model that can control any kind of robot body, even ones that are completely out-of-distribution. It can also adapt to large changes such as locked or damaged limbs even without training on them.  We hypothesize that the key to learning such controllers is to train on a wide task distribution with extreme domain randomization. When faced with large task diversity during training, policies must learn general strategies to efficiently perform task identification and then adapt control to achieve good task performance \cite{team2023human}. A similar phenomenon is observed in LLMs where web-scale pretraining leads to the ability to adapt in-context to new tasks at test time \cite{brown2020language}. We design a large space of procedurally generated robots containing bipeds and quadrupeds of the legged and wheeled variety and aggressively randomize the parameters of these robots including large variations in inertia, control gains and joint limits. 

Unfortunately, existing approaches are unable to learn effective policies in this task space. These only maintain a few hundred milliseconds of environment interaction in context. While this is sufficient when training specialist controllers on single embodiments, this short history doesn't contain enough information to adapt on the large task space. To increase the information available to the policy, we extend the context length by several orders of magnitude. We find that these simple changes, coupled with large-scale training, leads to emergent adaptation strategies.

We test \ours on unseen real-world robots including the Go2 (wheeled and legged variants) in both biped and quadruped mode, G1 and H1 robots among others. We find that in out-of-distribution (OOD) scenarios, \ours can improve performance across multiple episodes (Fig.~\ref{fig:adapt_motivation}). In particular, when testing with double the amount of domain randomization the policy was trained with we see that \ours can continuously adapt behavior across seconds of time (at least two orders of magnitude larger than prior work). This ability has at least two practical implications. First, \ours is more likely to bridge the sim2real gap better since it can withstand larger domain randomization at training time and adapt to OOD dynamics at test time. Second, since \ours can adapt to large changes in morphology, robots equipped with such a policy are likely to continue to function even in the face of adverse circumstances such as the loss of a limb or worse. 

\looseness=-1 Although we only present applications to locomotion in this paper, our method is simple and generally applicable in any scenario where large-scale data can be generated, such as in simulation. We envision this being useful for training generalist controllers for other low-level robotic skills as well. 

\section{Method}
\label{sec:method}
\begin{figure*}[t!]
    \centering
\includegraphics[width=\linewidth]{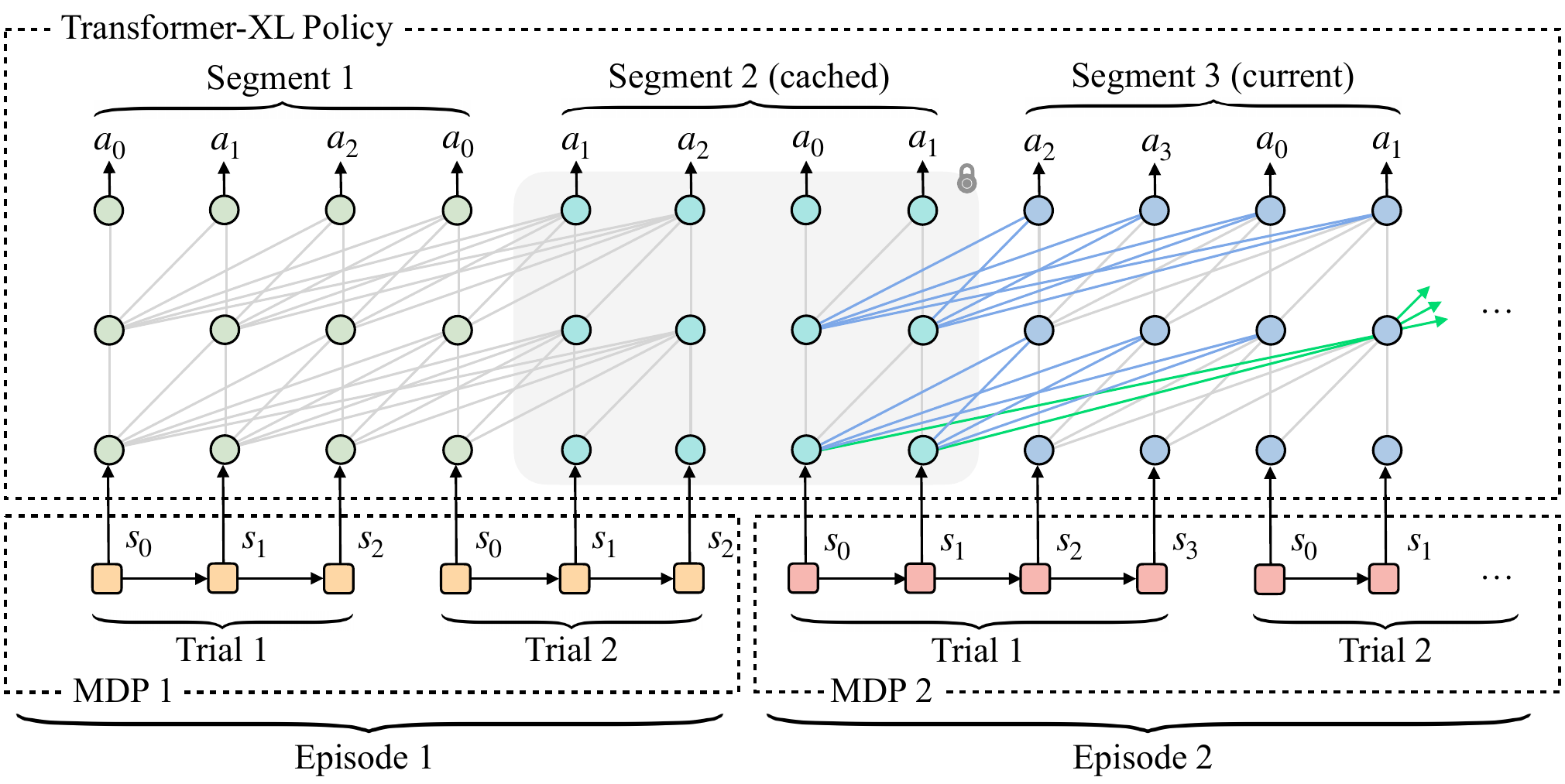}
    \caption{\small To enable long-context adaptation, we allow \ours to attend to states from prior trials. We use a long-context Transformer backbone, that divides context into fixed-length segments. When processing the current segment, keys and values are also computed from states in the previous (cached) segment as well, but gradients are not propagated (\textcolor[RGB]{100,100,255}{blue lines}). As the number of layers increases, this allows \ours to use information even segments prior to the cached one (\textcolor{green}{green lines}).}
    \label{fig:main_fig}
\end{figure*}

In this paper, our aim is to train a generalist locomotion policy that can robustly control any robot and adapt to with drastic changes. Given an unseen robot, the policy can be quickly deployed within seconds of adaptation. Our \ours, achieves the goal with a pre-training procedure that allows efficient in-context learning during test time (Sec. \ref{subsec:in_context}), and a policy architecture that enables long-term memory for adaptation (Sec. \ref{subsec:arch}). An overview of \ours is illustrated in Fig.~\ref{fig:main_fig}.

\subsection{Pre-training to Learn in Context}
\label{subsec:in_context}

Legged robots exhibit significant variations in their kinematics and dynamics, necessitating drastically different control strategies for them. One way to train a unified policy across diverse robots is to condition the policy on detailed kinematics and dynamics information~\cite{bohlinger2024one}. However, this method typically demands extensive system identification and remains susceptible to inaccuracies caused by wear and tear over time. In contrast, \ours takes a different approach by enabling in-context learning during test time. 

However, the naive approach of taking existing locomotion policy architectures and applying them to our procedurally generated task space doesn't work well. This is because most locomotion controllers are myopic and only use a few hundred milliseconds of history to adapt. This works well for a narrow task space, but with a large task space, this short history doesn't contain enough information to adapt effectively and leads to failure. Therefore, we need to train with much longer context lengths that sometimes may span trial boundaries \citep{duan2016rl}. Specifically, we let the policy run multiple trials in an episode. Memory persists across trials, and the objective is to maximize the expected cumulative reward across the entire episode.

\paragraph{Formulation}
\looseness=-1 Let $M = (\mathcal{S}, \mathcal{A}, \mathcal{T}, \mathcal{R}, \rho_0, \gamma, T)$ represent a Markov Decision Process (MDP), with state space $\mathcal{S}$, action space $\mathcal{A}$, transition function $\mathcal{T}: \mathcal{S} \times \mathcal{A} \rightarrow \mathcal{S}$, reward function $\mathcal{R}: \mathcal{S} \times \mathcal{A} \rightarrow \mathbb{R}$, initial state distribution $\rho_0$, discount factor $\gamma \in [0,1)$, and horizon $T$. A \emph{trial} is a single trajectory $\tau^M = [s_0, a_0, \dots]$ generated from initial state $s_0 \sim \rho_0(s_0)$. An \emph{episode} consists of $k$ trials, $\mathcal{E}^{M,k} = [\tau_1^M, \dots, \tau_k^M]$, where $k$ is sampled from $\rho_K = \text{Uniform}(1, \dots, K)$. A history-dependent policy $\pi_\theta(a_t \mid s_{0:t}, a_{0:t-1})$ selects actions based on the entire previous state and action history within an episode. Let $H_i$ denote the cumulative trajectory length up to trial $i$, with the convention $H_0=0$. Consider a distribution $\rho_{\mathcal{M}}$ over a set of MDPs $\mathcal{M}$. The objective is to maximize the expected discounted return:
\begin{equation}
J(\pi_\theta) = \mathop{\mathbb{E}}_{\substack{M \sim \rho_{\mathcal{M}} \\ k\sim \rho_K \\ \mathcal{E}^{M,k} \sim \pi_\theta }} \left[\sum_{i=1}^{k} \sum_{t=0}^{T} \gamma^{t + H_{i-1}} \mathcal{R}^M(s_t, a_t)\right].
\end{equation}

\looseness=-1 \paragraph{Task Generation} The above formulation assumes access to a distribution over MDPs. In our setting, we procedurally generate various legged robots and randomize their physical parameters to create diverse MDP instances. 
Specifically, we mainly focus on common morphologies such as bipeds, quadrupeds, and their wheeled counterparts without actually incorporating any exact robot parameters available on the market. For each morphology, we incorporate common design principles of existing legged robots. We randomize usual kinematic parameters, including joints, sequence of joints, etc. Dynamics parameters are randomized as well, including mass, center of mass, among other things.
See Fig.~\ref{fig:teaser} for a sample of generated robots and the appendix for more information. 

\paragraph{Unified Observation, Action and Reward Space} To train a policy shared across different legged robots, we use a simple \textit{unified joint space} for consistent observation and action representation by concatenating a superset of joints that subsumes the number of motors in the majority of contemporary legged robots.
The policy outputs target joint positions within the unified joint space, from which relevant joint states are extracted for each specific robot.
For rewards, we adopt a design scheme inspired by \cite{agarwal2023legged}, consisting of a command tracking reward and penalty terms to reduce torque usage, encourage smooth actions, and prevent hardware damage. 

\paragraph{Large-Scale RL in Simulation}
\looseness=-1 Locomotion tasks typically span long trials. To emphasize the adaptation learning process, we modify our formulation slightly: instead of directly sampling a fixed number of trials, we sample an adaptation time budget of $u$ seconds, $u \sim \text{Uniform}(0, U)$, allowing the policy to conduct an arbitrary number of trials within this allocated time before a final trial. Under the modified setting, we optimize the policy with Proximal Policy Optimization (PPO)~\cite{schulman2017proximal} at scale in physical simulation. Training proceeds in two phases: initially, we use shorter trial durations and smaller values of $U$ to prioritize the acquisition of adaptive behaviors. Subsequently, we extend trial lengths and adaptation time to prepare the policy for real-world deployment.

\paragraph{Deployment} A pre-trained policy can perform in-context learning in both zero-shot and few-shot manner. In zero-shot setting ($u=0$), the policy quickly adapts to an unseen embodiment within seconds, often achieving stable locomotion immediately. However, initial exploration may occasionally result in failures. In such cases, the policy benefits from few-shot learning, progressively improving performance by leveraging experiences from previous trials.

\subsection{Policy Architecture with Long-Term Memory}
\label{subsec:arch}

A Transformer-based policy has strong representational capacity and can attend to arbitrary positions in its history, essential for effective large-scale cross-embodied learning. However, high-frequency control tasks quickly generate extensive contexts—even short experiences lasting around 5 seconds can translate into hundreds of tokens, posing significant computational challenges for a vanilla Transformer where cost scales quadratically with sequence length. For computational tractability, we leverage the Transformer-XL (TXL) architecture~\cite{dai2019transformer} as the backbone of our history-dependent policy, which incorporates a segment-level recurrence mechanism to enable modeling of long-range dependencies.
We encode observations at timestep $t$ through an MLP into an embedding $x_t$. These embeddings serve as inputs to the TXL policy, whose outputs are subsequently decoded into actions and value estimates by separate actor and critic heads (each an MLP). Below, we detail the Transformer-XL backbone (Fig.~\ref{fig:main_fig}) and describe our approach to accelerating inference.

\paragraph{Transformer-XL  Policy} Suppose $\mathbf{o}_t$ are the observations at each timestep encoded into tokens $\mathbf{x}_t$. Transformer-XL (TXL) divides tokens into segments of length $L$, consider two such consecutive segments: $\sigma_z = [\mathbf{x}_{z,1},\dots,\mathbf{x}_{z,L}]$ and $\sigma_{z+1} = [\mathbf{x}_{z+1,1},\dots,\mathbf{x}_{z+1,L}]$. Let $\mathbf{h}_z^n \in \mathbb{R}^{L\times d}$ denote the hidden states of the $n$-th Transformer layer for segment $\sigma_z$. TXL uses cached hidden states from previous segments to compute current attention keys and values but doesn't backpropate gradients to them to save compute and memory. Formally, given previous hidden states $\mathbf{h}_{z}^{n-1}$, TXL computes:
\begin{align}
\tilde{\mathbf{h}}_{z+1}^{n-1} &= \left[\text{SG}(\mathbf{h}_{z}^{n-1}) \circ \mathbf{h}_{z+1}^{n-1}\right], \\
\mathbf{q}_{z+1}^{n},\ \mathbf{k}_{z+1}^{n},\ \mathbf{v}_{z+1}^{n} 
&= \mathbf{h}_{z+1}^{n-1} \mathbf{W}_q,\ 
   \tilde{\mathbf{h}}_{z+1}^{n-1} \mathbf{W}_k,\ 
   \tilde{\mathbf{h}}_{z+1}^{n-1} \mathbf{W}_v,\\
\mathbf{h}_{z+1}^{n} &= \text{Transformer-Layer}(\mathbf{q}_{z+1}^{n},\ \mathbf{k}_{z+1}^{n},\ \mathbf{v}_{z+1}^{n})
\end{align}
where $\text{SG}(\cdot)$ denotes a stop-gradient operation, $\circ$ indicates concatenation along the sequence length dimension, and $\mathbf{W}_q$, $\mathbf{W}_k$, $\mathbf{W}_v$ represent learnable model parameters. We use the standard Transformer architecture~\cite{vaswani2017attention} for TXL.

In RL training, each segment is cached to condition the subsequent iteration. This segment-level recurrence results in an effective context length growing as $O(NL)$ where $N$ is number of layers and $L$ is the segment length. For instance, a Transformer with 6 layers and segment length of 128 achieves a largest possible memory of 896 timesteps, corresponding to approximately 18 seconds at a control frequency of 50 Hz. Note that segments may span episode boundaries, in which case the attention mask is modified such that attention doesn't span episode boundaries.   

\paragraph{Inference Acceleration} The inference complexity of Transformer scales quadratically with segment length, which can significant slow down inference speed. To alleviate the computational bottleneck, we leverage KV-cache ~\cite{pope2023efficiently} to reduce inference complexity from quadratic to linear dependence on timestep $t$. During training-time rollouts, we initialize the KV-cache using keys and values derived from the previous segment's hidden states, and then append new timesteps incrementally. For real-time deployment, the KV-cache maintains only the most recent $2\times L - 1$ timesteps, which corresponds to the maximum directly conditioned history length encountered during training.
    

\section{Experimental Results}
\label{sec:result}



\subsection{Simulation Evaluation}
We compare the performance of \ours with several baselines in simulation. We additionally assess how the policy improves with growing adaptation time and examine the dynamics of its internal representations during adaptation.

\vspace{-0.5em}

\begin{table}[t!]
\centering
\resizebox{1.0\linewidth}{!}{%
\begin{tabular}{cccccccccccc}
\toprule
\multirow{2}{*}{\textbf{Method}} & \multicolumn{5}{c}{\textbf{Biped}} & \multicolumn{3}{c}{\textbf{Quadruped}} & \multicolumn{2}{c}{\textbf{Wheeled}} & \multirow{2}{*}{\textbf{Average}} \\\cmidrule{2-6}\cmidrule(lr){7-9}\cmidrule(lr){10-11} 
 & G1      & H1        & GR1       & TRON1     & BK-H  & A1 & Spot & AnyMal C & TRON1-W & Go2-W &  \\ \midrule
Ours (zero-shot)   & 0.96$\pm$0.16 & 0.98$\pm$0.12 & 0.95$\pm$0.21 & 0.87$\pm$0.31 & 0.98$\pm$0.11 & 0.92$\pm$0.26 & 1.00$\pm$0.06 & 0.95$\pm$0.22 & 0.99$\pm$0.07 & 0.97$\pm$0.16 & 0.96$\pm$0.19 \\
Ours (few-shot)       & 0.98$\pm$0.11 & 0.99$\pm$0.08 & 0.99$\pm$0.09 & 0.96$\pm$0.18 & 0.99$\pm$0.09 & 0.94$\pm$0.23 & 1.00$\pm$0.05 & 0.96$\pm$0.19 & 1.00$\pm$0.04 & 0.98$\pm$0.10 & 0.98$\pm$0.13 \\
GRU         & 0.09$\pm$0.06 & 0.08$\pm$0.11 & 0.09$\pm$0.11 & 0.03$\pm$0.03 & 0.47$\pm$0.40 & 0.46$\pm$0.44 & 0.68$\pm$0.37 & 0.42$\pm$0.45 & 0.66$\pm$0.34 & 0.74$\pm$0.36 & 0.37$\pm$0.41 \\
Conditioning   & 0.94$\pm$0.17 & 0.94$\pm$0.16 & 0.81$\pm$0.33 & 0.07$\pm$0.15 & 0.62$\pm$0.38 & 0.93$\pm$0.24 & 0.99$\pm$0.09 & 0.89$\pm$0.28 & 0.72$\pm$0.40 & 0.92$\pm$0.22 & 0.78$\pm$0.37 \\ \midrule
Expert Policy     & 1.00$\pm$0.00 & 1.00$\pm$0.00 & 1.00$\pm$0.00 & 1.00$\pm$0.00 & 1.00$\pm$0.00 & 0.97$\pm$0.18 & 0.99$\pm$0.08 & 1.00$\pm$0.04 & 1.00$\pm$0.00 & 0.98$\pm$0.12 & 0.99$\pm$0.07 \\
\bottomrule
\end{tabular}
}
\vspace{5pt}
\caption{
\small
We evaluate each method across 10 previously unseen robots spanning three categories in 1,000 simulation environments per robot, with rough terrains and extensive domain randomization. Each entry reports the average displacement towards a randomly sampled goal normalized to $[0, 1]$.
\ours performs competitively in zero-shot settings and shows further improvement with a brief 5-second adaptation window (10\% improvement on TRON1). On average, it approaches the performance of expert policy, outperforming both GRU and Conditioning baselines. 
}
\centering
\label{table:results}
\vspace{-1.5em}
\end{table}

\paragraph{Baselines} We consider the following baselines 
(1) \textit{GRU Policy} A GRU-based policy trained under the same setting as \ours to evaluate the effect of architectural choice;
(2) \textit{Conditioning} A Transformer policy conditioned on privileged kinematics information but with restricted memory for adaptation (context length 64);
(3) \textit{Expert Policy} A Transformer policy trained directly on the target embodiment, serving as an upper-bound with full access to the deployment domain.
\ours is evaluated in both zero-shot and few-shot settings. Under the few-shot setting, the policy is allowed a 5-second adaptation window prior to evaluation.

\vspace{-0.5em}

\paragraph{Evaluation on Unseen Robots} 
We test all methods on a set of previously unseen robots: five bipeds (Unitree G1, H1, Fourier GR1, LimX Dynamics TRON1, Berkeley Humanoid BK-H), three quadrupeds (Unitree A1, Boston Dynamics Spot, ETH AnyMal C), and two wheeled robots (TRON1-W, Untiree Go2-W). Evaluation is conducted across 1,000 randomized simulation environments with rough terrains and intensified domain randomization.
We randomly sample a goal in each environment an measure the average displacement towards the goal. Results are presented in Tab.~\ref{table:results}. 
Despite having no prior access to these robots, our method achieves performance close to the expert policies in both zero-shot and few-shot settings. While it adapts effectively in a zero-shot manner, extended adaptation proves beneficial in challenging scenarios. For example, on TRON1—a biped without ankle joints—5 seconds of pre-evaluation adaptation yields a 10\% performance gain.
By contrast, the GRU baseline performs poorly with an average score 0.37, highlighting the importance of architecture choice for cross-embodiment generalization. The Conditioning baseline benefits from morphology inputs but is constrained by its short context window, resulting in moderate performance.

\vspace{-0.5em}

\begin{figure*} [t!]
\centering
    \subfloat[\label{fig:adapt_surv}]
    {
        \includegraphics[width=0.45\linewidth]{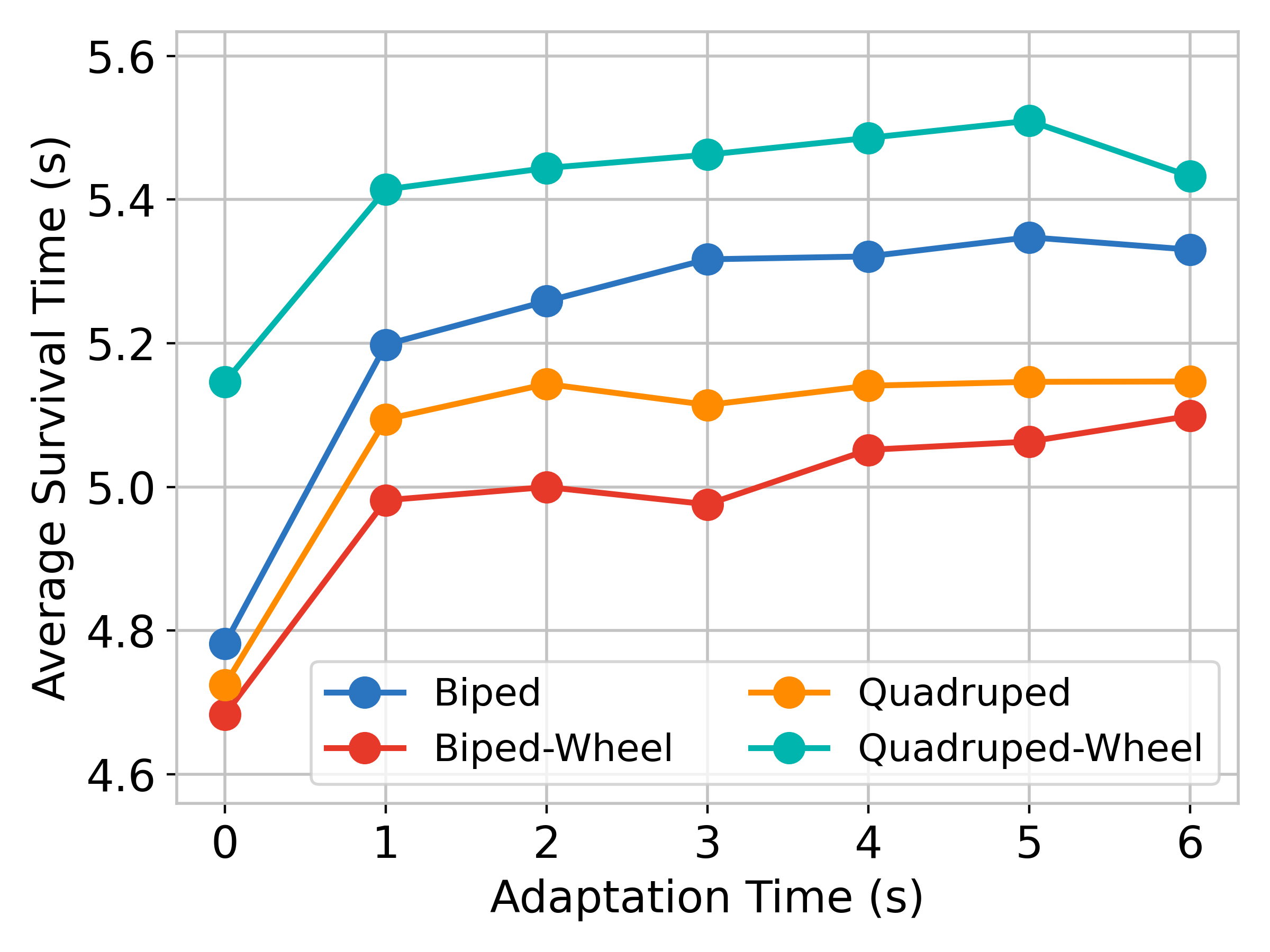}
    }
    \subfloat[\label{fig:adapt_repr}]
    {
        \includegraphics[width=0.45\linewidth]{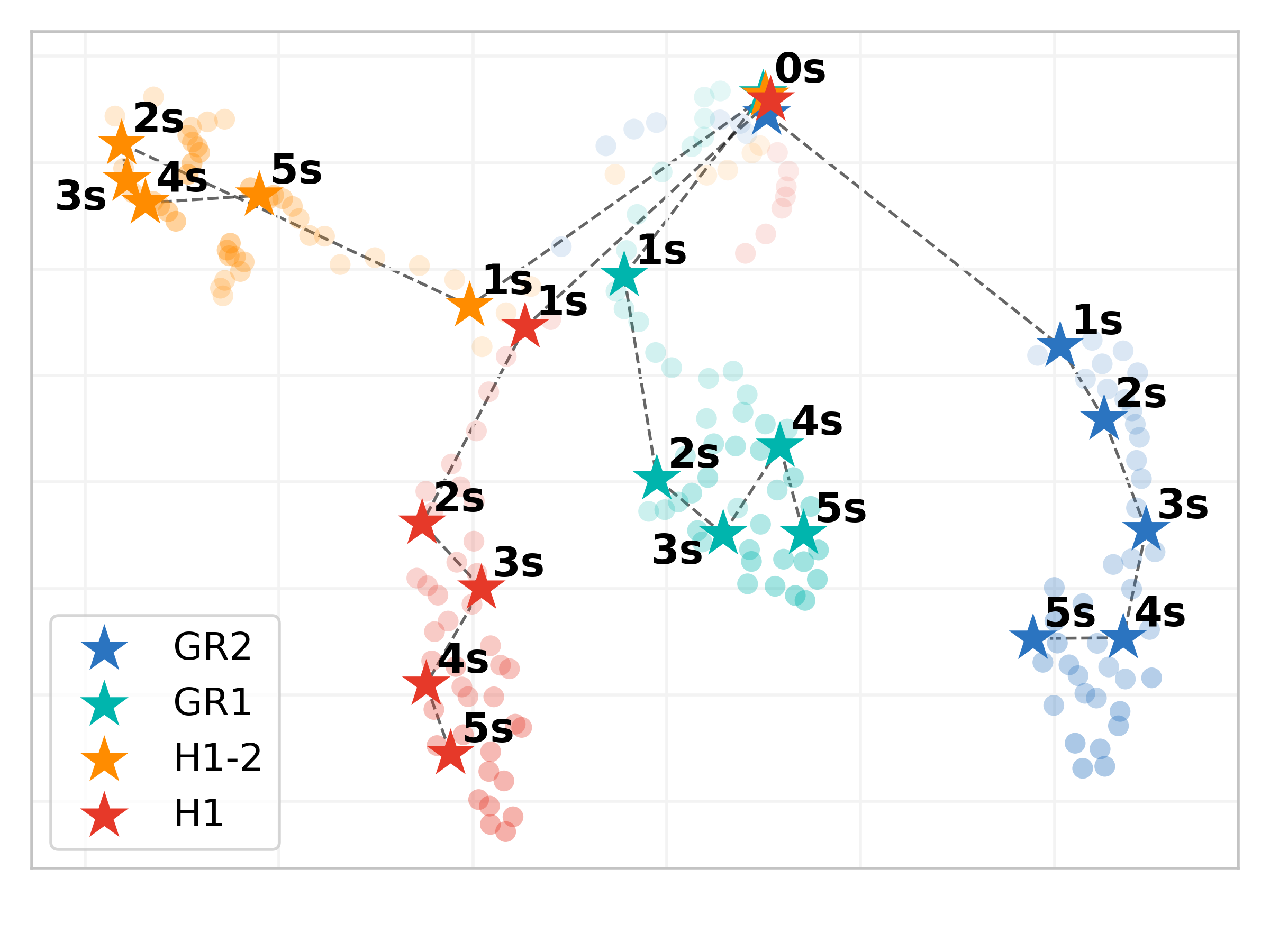}
    }
    \caption{
    \small
    \textbf{(a)} Performance of \ours with different adaptation time. 
    We deploy the policy on procedurally generated robots across four morphology types under intensified domain randomization, and report average survival over 6-second rollouts across varying adaptation time budgets. Our policy consistently benefits from longer adaptation.
    \textbf{(b)} Temporal evolution of policy representations. 
    We visualize the mean output of the second Transformer layer over time across 4,096 zero-shot rollouts of four humanoid variants. Representations begin similarly but diverge progressively, forming distinct clusters by 5 seconds. This indicates that \ours dynamically builds stable, embodiment-specific representations online, without explicit morphology specification.
    }
    \label{fig:adaptation} 
    \vspace{-1.2em}
\end{figure*}

\paragraph{Adaptation Performance}
To assess adaptation capability of \ours, we double domain randomization intensity and deploy procedurally generated robots on rough terrain. We record the average time to fall over 6-second rollouts across varying adaptation time budgets (50K episodes per setting). As shown in Fig.~\ref{fig:adapt_surv}, \ours can adapt to challenging domains effectively even zero-shot, with substantial gains from a short adaptation period. All morphologies exhibit improved survival with increased adaptation time, highlighting the advantage of long-context learning.
\vspace{-0.5em}

\paragraph{Representation Dynamics}
We further analyze adaptation by tracking the temporal evolution of internal representations. Specifically, we monitor the average output of the second Transformer layer over 4,096 zero-shot rollouts of four humanoid variants (Fig.~\ref{fig:adapt_repr}). Initially similar embeddings (within the first second) diverge over time, reflecting the policy’s growing specialization on different embodiments. By 5 seconds, distinct representation clusters emerge, indicating the policy’s ability to form stable, embodiment-specific internal models without explicit robot descriptors.

\subsection{Real-World Evaluation}
We first evaluate how \ours transfers to unseen robots in real world. Our test set includes four Unitree robots: G1, H1, Go2, and Go2-W, as illustrated in Fig.~\ref{fig:teaser}. Notably, we also evaluate Go2 and Go2-W in a bipedal mode by activating only their rear legs. Our policy is deployed zero-shot on these robots and adapts within seconds, without any prior tuning or retraining.

\begin{figure*} [t!]
    \centering
    \includegraphics[width=\linewidth]{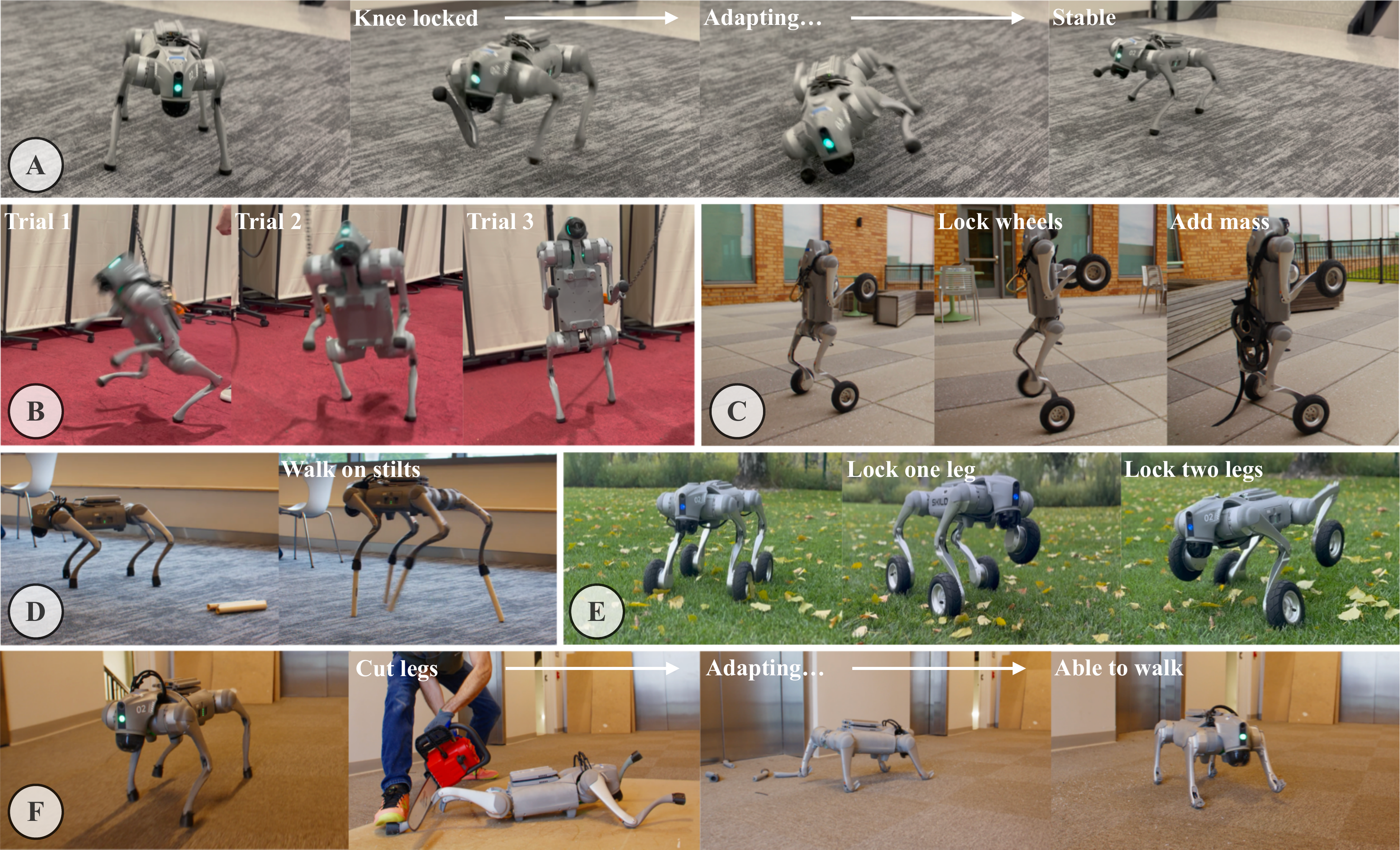}
    \vspace{-1.3em}
    \caption{
        \looseness=-1
        \small
        We showcase emergent adaptation behaviors, all from the same \ours model. 
        \textbf{A}: A knee is locked, and \ours adapts to walk with three remaining legs. This is not a part of training. 
        \textbf{B}: If \ours is unable to adapt within a trial, it retains memory of previous trials and improves performance. 
        \textbf{C}: The wheels are locked and \ours detects the sudden change, and shifting to a walking gait. \ours also adapts when additional mass is attached. 
        \textbf{D}: Stilts are attached, and \ours adapts to the extended leg length to continue walking.
        \textbf{E}: With one or two legs locked, \ours adapts its motion to maintain balance.
        \textbf{F}: After lower legs are cut, the robot adapts after tens of seconds to walk on its knees despite not seeing it at training time. 
    }
    \label{fig:real_adapt} 
    \vspace{-1.7em}
\end{figure*}

\ours is trained at scale across diverse morphologies under heavy online perturbations , and it incorporates memory across trials. We hypothesize that this enables the policy to handle large real-world disturbances such as sudden morphology changes and added payloads, and to leverage prior experience for efficient control. We showcase these abilities below (see Fig.~\ref{fig:real_adapt}).

\vspace{-0.2em}
\looseness=-1
\textbf{A, E: Leg locking}
While Go2 is walking normally, a knee is locked by overriding the PD setpoint to a hardcoded value. This converts the quadruped into a three legged robot, destabilizing it. To recover, the policy must adopt a gait that keeps the center of mass within the triangular support polygon—an unseen scenario since three-legged robots are outside our design space. \ours initially tips forwards but learns to shift its weight backward onto three legs and is even able to walk after 2-3s of adaptation. Similar behavior occurs on a wheeled quadruped: when one or two legs are locked, \ours adapts its gait to redistribute load, preserving balance and mobility despite reduced actuation.

\vspace{-0.2em}
\looseness=-1
\textbf{B: Adaptation across trials}
In highly unstable cases, adaptation within a single attempt may be impossible. We run \ours as a biped with no ankle motor and a single point of support. This is a highly unstable platform and affords very little time to stabilize before failure occurs. Since \ours has no idea about the morphology it fails in the first trial. However, we can retain the failed trajectory information in the TransformerXL cache and attend to it leading to improved performance in subsequent trials. On the third trial, the robot walks stably in biped mode and is even robust to pushes and weights.

\vspace{-0.2em}
\looseness=-1
\textbf{C: Wheel locking and payload change}
While Go2-W is initially rolling, we programmatically lock the wheels during execution by setting the gains to zero in software. This instantly alters the robot’s dynamics, as it can no longer use rolling to locomote. \ours detects this discrepancy, as sending commands to the wheel no longer has the same effect. It quickly adapts to this change and switches to a walking gait instead like in a standard legged biped. At some point the wheels are again unlocked. \ours  detects this and switches to a lower energy rolling gait. Similarly, when external mass is added, \ours adjusts its gait to maintain balance.

\vspace{-0.2em}
\looseness=-1
\textbf{D: Walking on stilts}
We attach stilts to the robot’s legs, increasing the effective leg-to-body length ratio far beyond those seen during training. This alters both kinematics and dynamics of locomotion by raising the center of mass and changing swing/stance leverage. Initially, the robot takes a few unstable steps, but \ours quickly adapts its coordination—adjusting step timing and foot placement—to account for the extended morphology and is able to walk forward with stable, repeatable gaits.

\vspace{-0.2em}
\looseness=-1
\textbf{F: Cutting lower legs} We cut the calf of the robot to its thigh. This essentially removes 4 degrees of freedom and lowers the limb length. This configuration is unseen at training. Initially, the robot is unable to locomote effectively and merely steps in place. However, after 7-8s of adaptation, it discovers that large amplitude swings are required at the thigh joint and is able to locomote effectively thereafter.

\section{Related Work}
\label{sec:related_work}
\vspace{-0.2em}

\subsection{Legged Locomotion} 
\vspace{-0.6em}
In recent years, sim2real transfer has emerged as an effective paradigm for training locomotion. For blind robots, this is done by training an adaptive policy in simulation with domain randomization and then transferring zero-shot to the real world, both for quadrupeds \citep{kumar2021rma, lee2020learning, nahrendra2023dreamwaq, margolis2023walk, 10530429, smith2023learning} and humanoids \citep{radosavovic2024real, cheng2024expressive, he2024hover, chen2024learning, xue2025unified, siekmann2021sim}. Subsequent works couple locomotion with vision to adapt walking behavior based on upcoming terrain geometry using either implicit \citep{agarwal2023legged, duan2024learning, cheng2024extreme, zhuang2023robot, luo2024pie, yang2023neural} or explicit mapping \citep{sun2025learning, miki2022learning, wang2025beamdojo}. In all these works, the controller is tuned for a specific robot with narrow dynamics randomization and a short context length (hundred millisecond) for adaptation. In contrast, we train a single, generalist policy that adapts over long contexts to vastly different embodiments and dynamics. 

\subsection{Cross-embodiment Learning} 
\vspace{-0.6em}

\looseness=-1 The success of large-scale pretraining in language and vision has spurred interest in developing similar models for robots that can absorb data from multiple embodiments. For locomotion, many methods focus on adding structural biases to the policy architecture based on the robot morphology in the form of message passing on graph neural networks \citep{huang2020one} or through attention masks in a transformer \citep{sferrazza2024body}. \citet{gupta2021embodied} co-evolve locomotion controllers with robot morphologies. Other methods focus on parameterizing the action and observation spaces of the policy to promote knowledge transfer. \citet{doshi2024scaling} train a transformer with imitation learning that uses different action heads to decode readout tokens from a transformer to different morphologies. \citet{bohlinger2024one} trains with RL on a few chosen robots with a custom latent observation space that encodes joint observations as well as precise kinematic structure of the robot. In contrast, \citet{reed2022generalist} simply trains with next token prediction on data from multiple embodiments. Closely related to our work, \citet{feng2023genloco} train a single controller across procedurally generated quadrupeds. The key difference from these works is that we train at massive scale with long contexts across a much larger task space consisting of wheeled and legged quadrupeds and humanoids. We show that this leads to novel and highly flexible adaptation strategies.  

\subsection{Adaptation}
\vspace{-0.6em}

\looseness=-1 Adaptation, meta-learning \citep{finn2017model} and in-context learning \citep{brown2020language} are related concepts which deal with the problem of achieving high task performance from a few examples. Meta-RL or fine-tuning can be used for rapidly adapting specialist locomotion policies to changes in dynamics \citep{bao2025toward, song2020rapidly, smith2022walk}. In game-like environments, \cite{duan2016rl} show that training with contexts spanning episodes leads to adaptive behavior where policies learn to improve from early mistakes. \citet{team2023human} shows that large-scale training in an open-ended task space leads to agents that learn to adapt on human-timescales by reasoning across long contexts. In our paper, we study adaptation in the high-frequency closed loop control setting, and present a single policy that adapts to varied embodiments and dynamics.


\section{Conclusion}
\looseness=-1 In this paper, we present \ours, a generalist locomotion policy that can robustly control a variety of legged and wheeled robots in the real world. This is enabled by emergent adaptive behavior where the policy can use long contexts as information to improve performance. Although we only discuss locomotion in this paper, we believe this recipe is general and can be used in the future to train generalist controllers for other skills, especially in simulation where such large-scale RL is possible.  
\label{sec:conclusion}


\section{Limitations}
\label{sec:limitation}
We identify several key limitations of our approach:
\begin{itemize}[noitemsep,topsep=0pt]
    \item Training \ours is highly resource-intensive compared to a specialist policy. While this is not surprising given the much wider task distribution and context, there may be algorithmic improvements possible in the massively parallel RL setting \cite{singla2024sapg}. 
    \item \ours relies on a procedurally generated task space which is hand-crafted. This might be hard to design in general. For future work, automated task generation from LLMs or other web-scale sources might be a more scalable approach. 
\end{itemize}

\subsection*{Acknowledgment}
We thank Kevin Gmelin, Jayesh Singla, Jared Meija, Shikhar Bahl for fruitful discussions. 


\clearpage


\bibliography{reference}  

\appendix

\part*{Appendix}

\section{Implementation Details}

This section outlines the procedures for generating diverse robot morphologies, randomizing environment dynamics, designing rewards, and training the \ours policy. 

\subsection{Procedural Robot Generation}

\begin{figure*} [h!]
\centering
    \subfloat[\label{fig:adapt_surv}]
    {
        \includegraphics[width=1.0\linewidth]{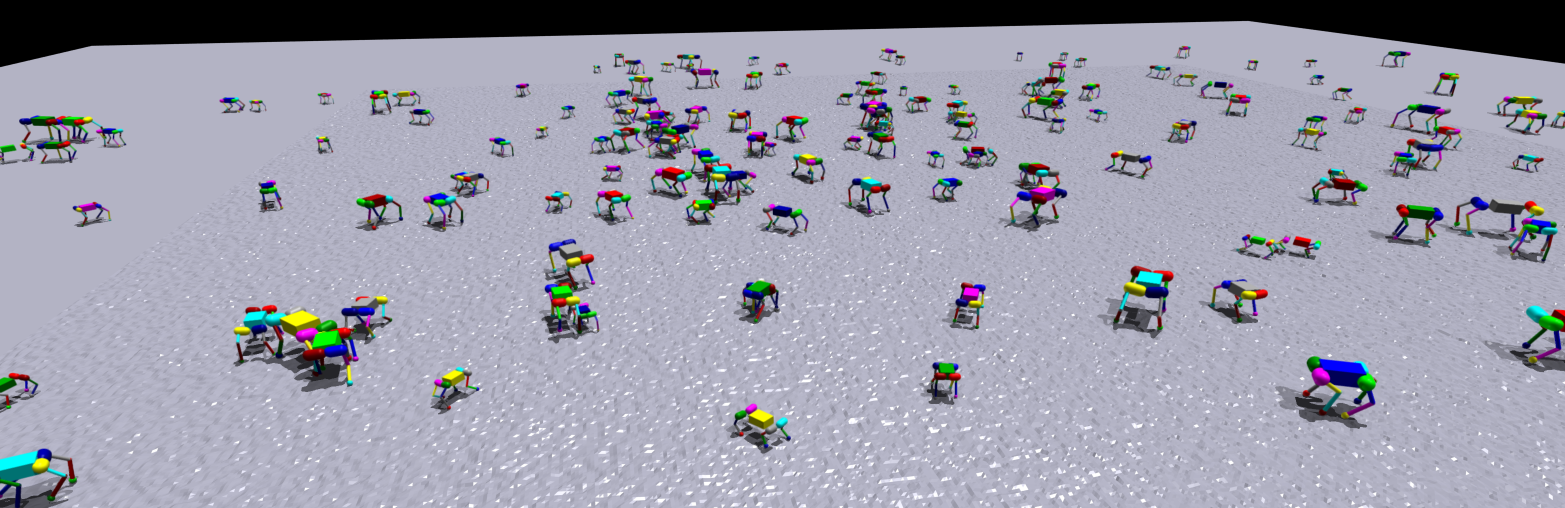}
    }
    \\
    \subfloat[\label{fig:adapt_repr}]
    {
        \includegraphics[width=1.0\linewidth]{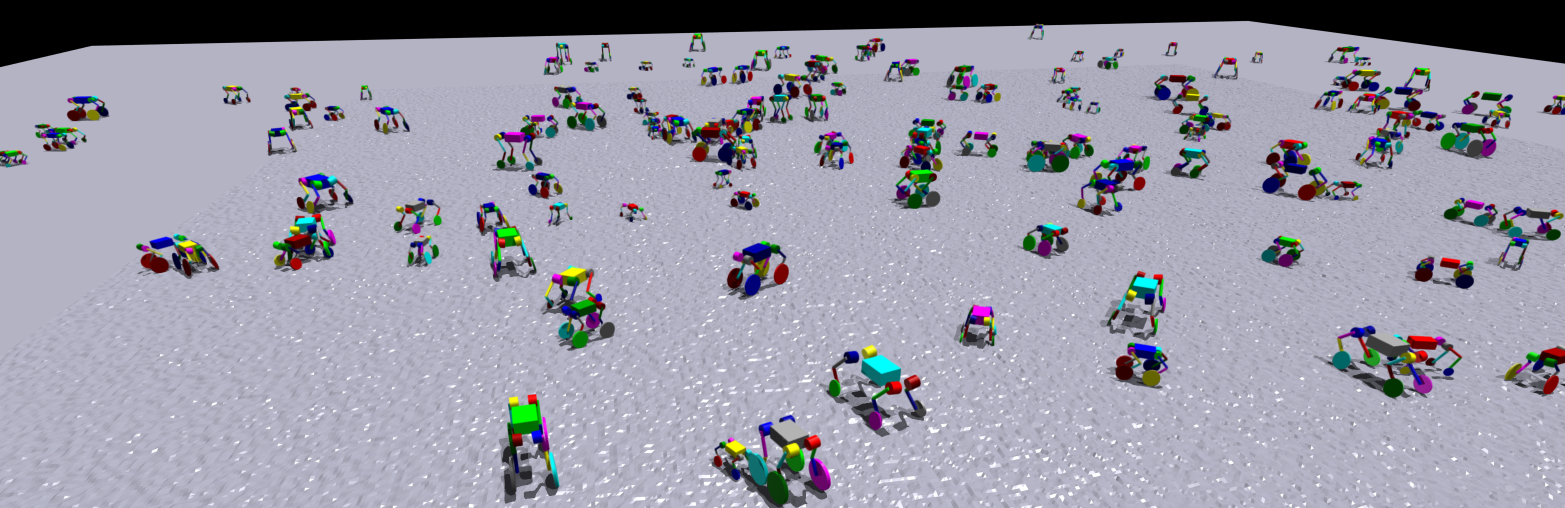}
    }
    \\
    \subfloat[\label{fig:adapt_surv}]
    {
        \includegraphics[width=1.0\linewidth]{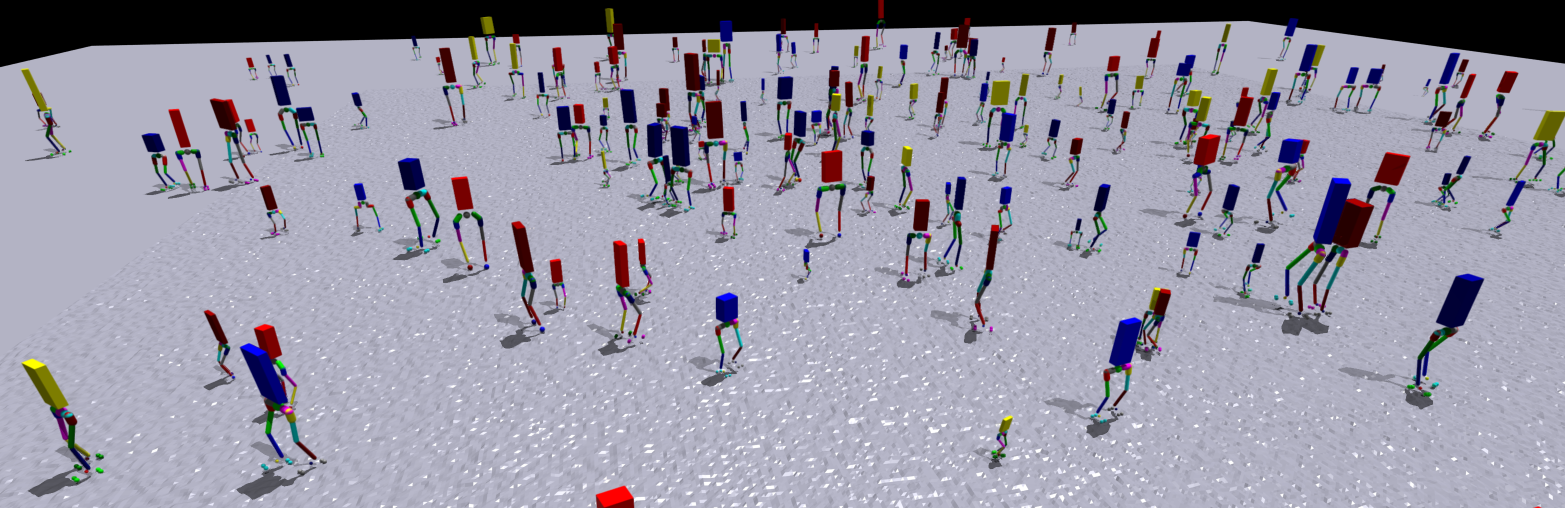}
    }
    \\
    \subfloat[\label{fig:adapt_repr}]
    {
        \includegraphics[width=1.0\linewidth]{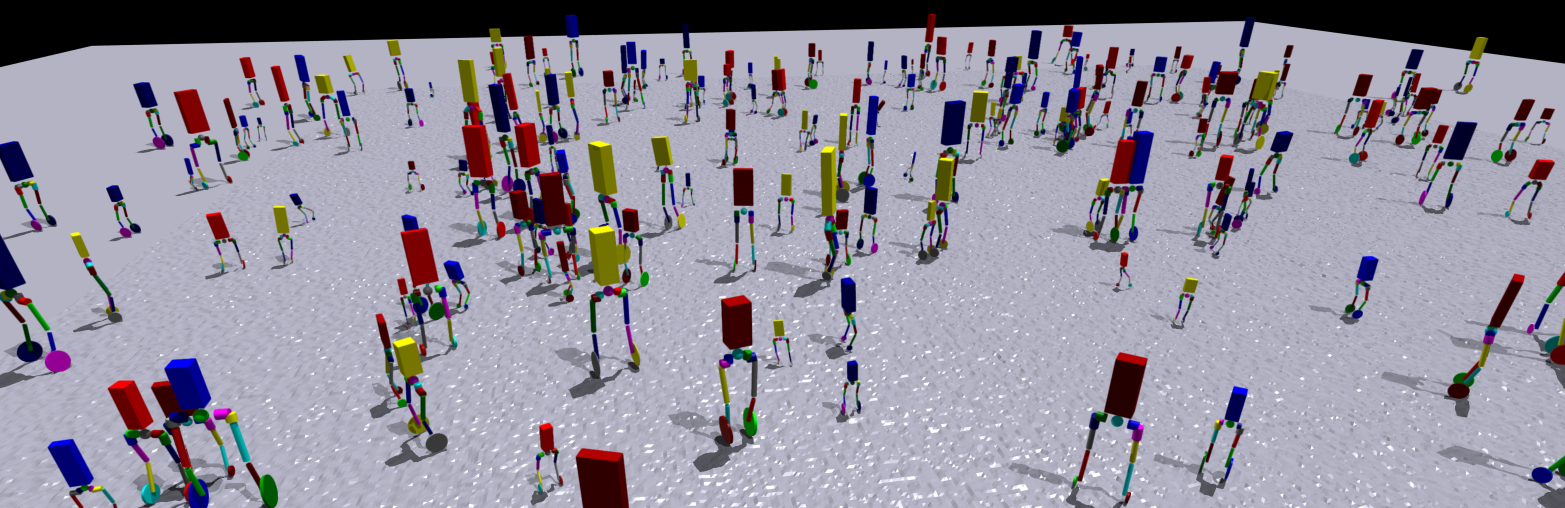}
    }
    \caption{
    \small
    Visualization of sample procedurally generated robots.
    \textbf{(a)} Quadrupeds. \textbf{(b)} Quadrupeds with wheels. \textbf{(c)} Bipeds. \textbf{(d)} Bipeds with wheels.
    }
    \label{fig:sim_training} 
\end{figure*}

We generate four categories of morphologies: quadrupeds, bipeds, and their respective wheeled counterparts. Random samples are illustrated in Fig.~\ref{fig:sim_training}. While our procedural generation covers a wide morphological space, it is not exhaustive. Our aim is to train a policy capable of generalizing to out-of-distribution robots. In practice, unseen robots often deviate from our design assumptions. For instance, Berkeley-Humanoid exhibits anhedral angles in both hip roll and yaw joints—our bipeds include at most one. Additionally, robots like G1 have joint axes offset along the $x$-axis, a feature not captured in our parameterization. As for dynamics randomization, we randomized the standard parameters~\cite{agarwal2023legged} but over a much wider range so that the model learns to adapt.

\subsection{Reward Design}
We design a unified reward structure combining morphology-agnostic and morphology-specific terms. For clarity, the time subscript $t$ is omitted in some expressions.

\textbf{Shared Reward Components}
\begin{itemize}
    \item \textbf{Linear velocity command tracking} $\exp(\|v_{xy} - v^{cmd}_{xy}\|^2/s_1)$
    \item \textbf{Angular velocity command tracking} $\exp(\|w_{z} - w^{cmd}_{z}\|^2/s_2)$
    \item \textbf{Base linear velocity penalty} $\|v_z\|^2$
    \item \textbf{Base angular velocity penalty} $\|w_{xy}\|^2$
    \item \textbf{Base height penalty} $\|x_z - x_z^{nominal} \|^2$ 
    \item \textbf{Joint acceleration penalty} $\|\ddot q\|^2$
    \item \textbf{Torque penalty} $-\|\tau\|^2$
    \item \textbf{Alive reward}: $1$
\end{itemize}

\section{Experiment Details}

\begin{figure*}[ht!]  
    \centering
    \includegraphics[width=1.0\textwidth]{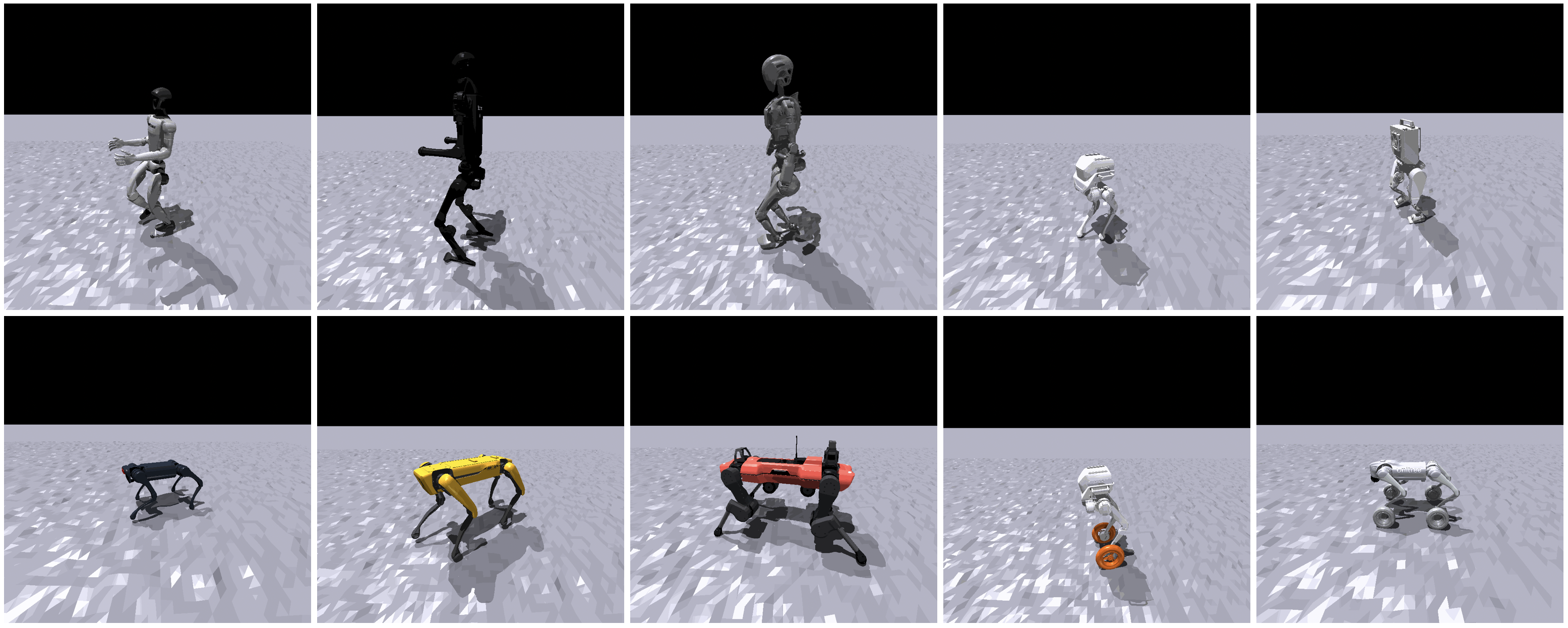}
    \vspace{-15pt}
    \caption{\small Robots where we compare with the baseline methods in simulation. First row: Unitree G1, H1, Fourier GR1, LimX Dynamics TRON1, Berkeley Humanoid. Second row: Unitree A1, Boston Dynamics Spot, ETH AnyMal C, LimX Dynamics TRON1-WF, Unitree Go2-W. }
    \label{fig:eval_robots}
\end{figure*}

\begin{figure*}[ht!]  
    \centering
    \includegraphics[width=1.0\textwidth]{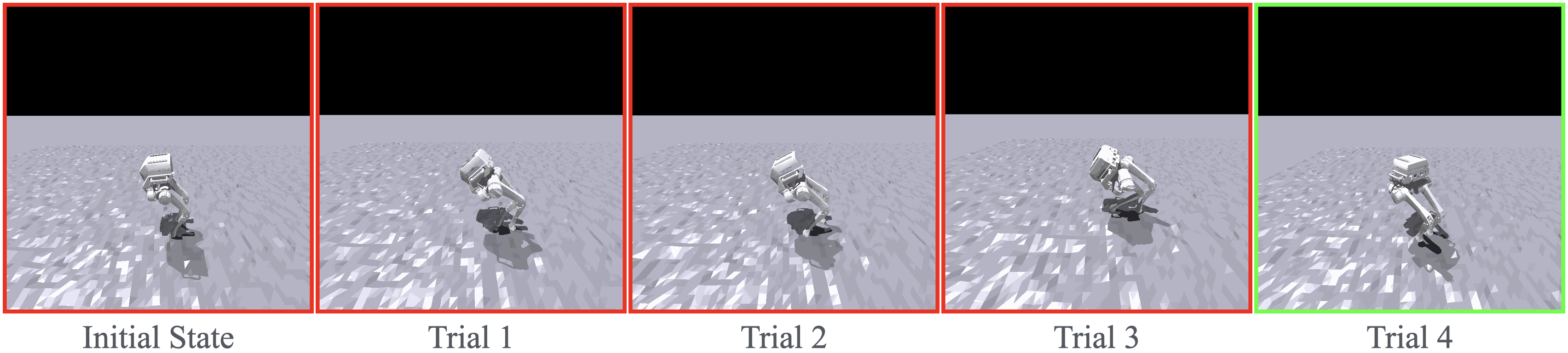}
    \vspace{-15pt}
    \caption{\small Cross-trial adaptation of TRON1 in simulation. The robot fails in early trials but progressively learns to stabilize and locomote effectively by Trial 4.
    }
    \label{fig:eval_adapt}
\end{figure*}

Fig.~\ref{fig:eval_robots} shows the 10 simulated robots used for comparison against baseline methods. Among them, we observe a notable sign of cross-trial adaptation in the TRON1 robot, which achieves a 10\% performance improvement through 5-seconds pre-evaluation adaptation. We find the robot initially fails in early trials but gradually learns to stabilize and control itself effectively. An illustrative example of this adaptation process is shown in Fig.~\ref{fig:eval_adapt}.

We provide videos of real world evaluation on \href{https://generalist-locomotion.github.io}{generalist-locomotion.github.io}.

\section{Computation Cost}
\ours is a highly performant generalist locomotion model trained with a high capacity Transformer model across 100k robots. Compared to a specialist policy, \ours takes ~500 times more compute for 100000 more robots -- this is a significant reduction in amortized compute cost per robot  (1 vs. 0.005 day).

\begin{figure}[h!]
    \centering
    \includegraphics[width=0.8\linewidth]{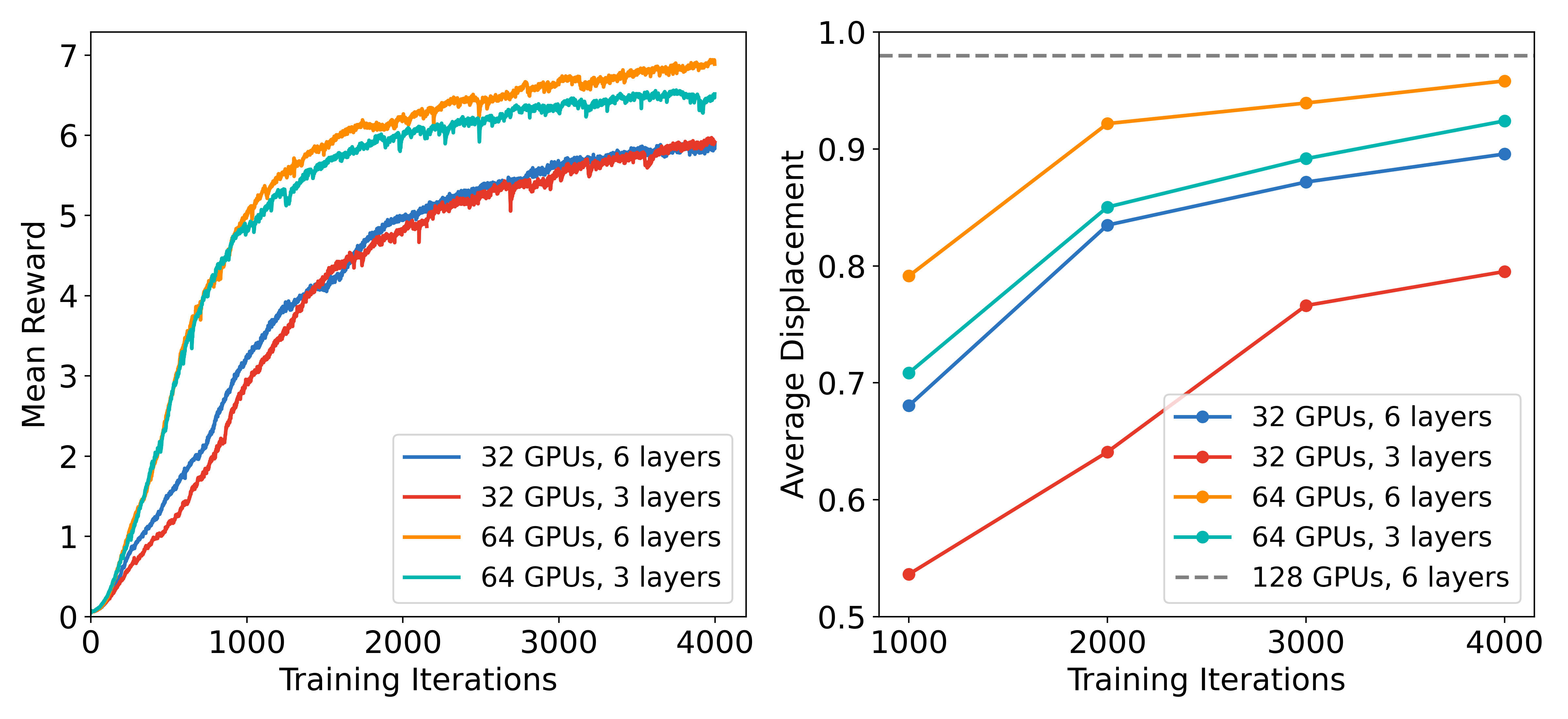}
    \caption{
    We train with different numbers of GPUs and TXL layers for 4K iterations (28 hours for 6-layer TXL), and evaluate on 10 robots. The left plot shows reward on robots in the train set, while the right reports avg. displacement of selected checkpoints on 10 eval robots (including OOD cases as described in Tab.~1). While more GPUs yields the best model, there is a favorable tradeoff between performance and compute.
    }
    \label{fig:computation}
\end{figure}

\paragraph{Optimizing compute further:} We trained \ours at scale to get the best possible model. However, there is a favorable trade-off between performance and compute -- even a variant trained with much lesser compute shows clear signs of adaptation and cross-embodiment generalization (Fig.~\ref{fig:computation}). Since we use a Transformer backbone, techniques from LLM literature can be directly used to reduce compute requirement further, such as, \href{https://arxiv.org/abs/1710.03740}{mixed precision}, \href{https://arxiv.org/abs/1910.02054}{ZeRO memory optimization}.

\end{document}